\documentclass[10pt,twocolumn,letterpaper]{article}

\usepackage{cvpr}              

\usepackage{graphicx}
\usepackage{amsmath}
\usepackage{amssymb}
\usepackage{amsthm}
\usepackage{amsfonts}
\usepackage{dsfont}
\usepackage{booktabs}

\usepackage{times}
\usepackage{amsfonts}
\usepackage{bm}
\usepackage{tabu}
\usepackage{multirow}
\usepackage{color}
\usepackage[abs]{overpic}
\usepackage{comment}
\usepackage{textcomp}
\usepackage[accsupp]{axessibility}

\definecolor{mypink}{RGB}{236, 2, 141}

\usepackage[capitalize]{cleveref}
\crefname{section}{Sec.}{Secs.}
\Crefname{section}{Section}{Sections}
\Crefname{table}{Table}{Tables}
\crefname{table}{Tab.}{Tabs.}

\newcommand*{\affaddr}[1]{#1} 
\newcommand*{\affmark}[1][*]{\textsuperscript{#1}}
\newcommand*{\email}[1]{\small{\texttt{#1}}}

\begin{document}

\title{Global Tracking via Ensemble of Local Trackers}

\author{Zikun Zhou\affmark[1,*], Jianqiu Chen\affmark[1,*], Wenjie Pei\affmark[1,\dag], Kaige Mao\affmark[1], Hongpeng Wang\affmark[1,2], and Zhenyu He\affmark[1,\dag]\\\affaddr{\affmark[1]Harbin Institute of Technology, Shenzhen}\quad\affaddr{\affmark[2]Peng Cheng Laboratory}\\
\email{zhouzikunhit@gmail.com\quad jianqiuer@gmail.com\quad wenjiecoder@outlook.com}\\ \email{maokaige.hit@gmail.com\quad wanghp@hit.edu.cn\quad zhenyuhe@hit.edu.cn}\\
}

\maketitle
\renewcommand{\thefootnote}{\fnsymbol{footnote}} 
\footnotetext[1]{Equal contribution.}
\footnotetext[2]{Corresponding authors.}

\begin{abstract}
The crux of long-term tracking lies in the difficulty of tracking the target with discontinuous moving caused by out-of-view or occlusion. Existing long-term tracking methods follow two typical strategies. The first strategy employs a local tracker to perform smooth tracking and uses another re-detector to detect the target when the target is lost. While it can exploit the temporal context like historical appearances and locations of the target, a potential limitation of such strategy is that the local tracker tends to misidentify a nearby distractor as the target instead of activating the re-detector when the real target is out of view. The other long-term tracking strategy tracks the target in the entire image globally instead of local tracking based on the previous tracking results. Unfortunately, such global tracking strategy cannot leverage the temporal context effectively.
In this work, we combine the advantages of both strategies: tracking the target in a global view while exploiting the temporal context. Specifically, we perform global tracking via ensemble of local trackers spreading the full image. The smooth moving of the target can be handled steadily by one local tracker. When the local tracker accidentally loses the target due to suddenly discontinuous moving, another local tracker close to the target is then activated and can readily take over the tracking to locate the target. While the activated local tracker performs tracking locally by leveraging the temporal context, the ensemble of local trackers renders our model the global view for tracking. Extensive experiments on six datasets demonstrate that our method performs favorably against state-of-the-art algorithms.
\end{abstract}

\section{Introduction}
The long-term visual tracking task has attracted more attention from the visual tracking community in recent years. Compared with short-term tracking, the long-term tracking task is much closer to the real-world application, due to the following two differences.
First, the average duration of the sequences in the long-term tracking benchmarks (such as LaSOT~\cite{LaSOT}, TLP~\cite{TLP}, and OxUvA~\cite{OxUvA}) is hundreds of seconds, which is much longer than that (tens of seconds) of the short-term tracking benchmarks (OTB2015~\cite{OTB2015}, TrackingNet~\cite{TrackingNet}, and GOT-10k~\cite{Got-10k}, to name a few).
Second, the long-term tracking task requires the algorithms to be able to cope with the discontinuous moving of the target caused by the disappearance and reappearance of the target.

\begin{figure}[t]
\centering
	\includegraphics[width=1.0\columnwidth]{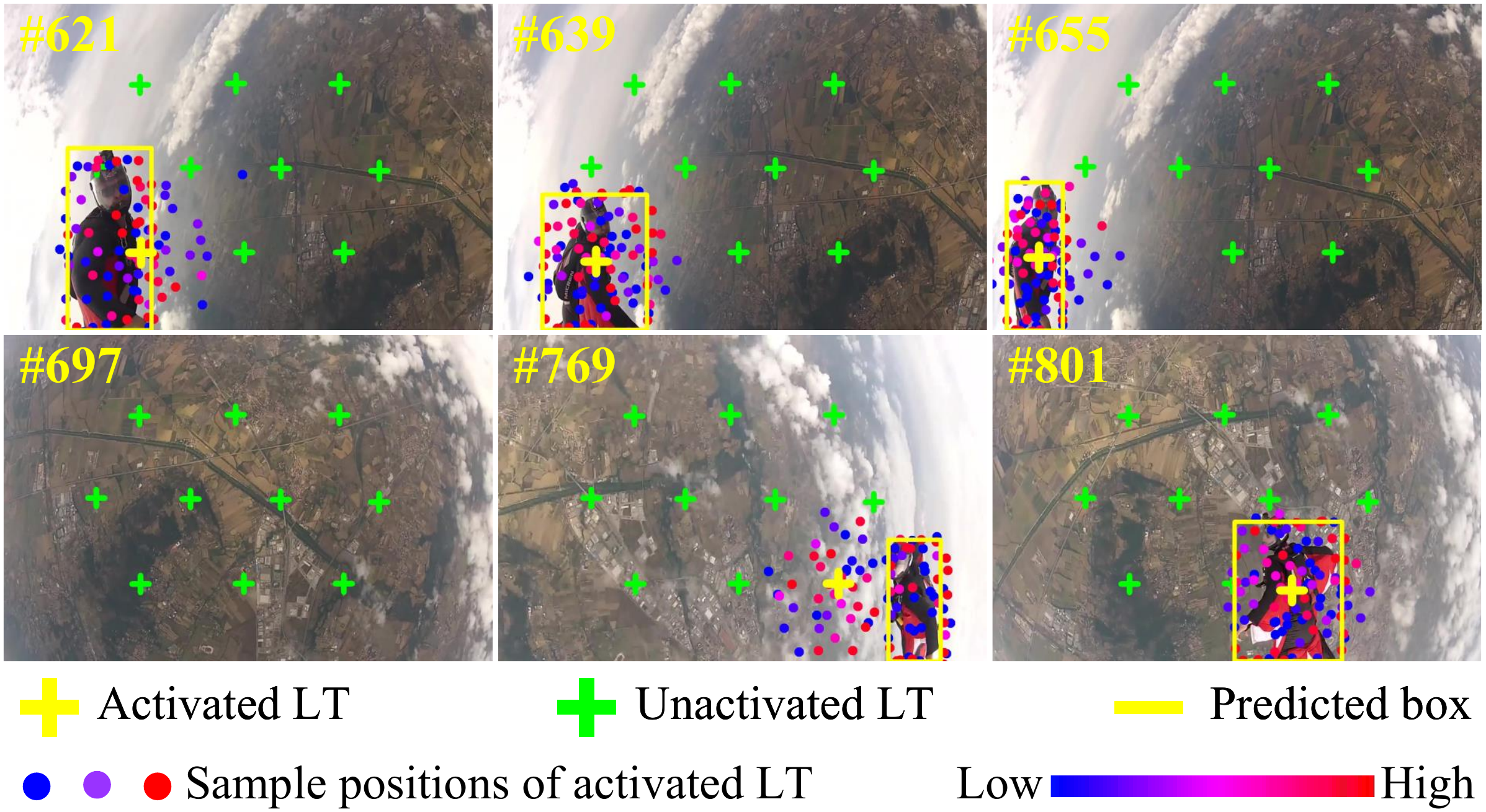}
	\caption{Illustration of the tracking process via ensemble of local trackers for tracking a wingsuit flyer, which disappears from the left-bottom corner of the view and then reappears from the right-bottom corner. The color of sample positions denotes the attention weight. At first, the Local Tracker (LT) at the left-bottom corner is activated and moves following the target to keep tracking it until the target disappears. While the activated local tracker loses the target, it moves back to where it starts moving, as shown in the $697^{th}$ frame. When the target reappears, the local tracker at the right-bottom corner is activated and keeps tracking the target. The ensemble of local trackers collaborates to achieve global tracking.}
	\vspace{-5mm}
	\label{Fig:Introduction}
\end{figure}

Most existing trackers~\cite{ECO,SiamFC,SiamRPN++,DiMP,SAOT} search for the target within a local image region, named local trackers, and thus cannot handle the frequently discontinuous moving of the target in the long-term tracking task. To handle this issue, a typical strategy~\cite{TLD,LCT,SPLT,LTMU} is to equip the local tracker with a global re-detector to detect the target after the local tracker fails. Such a strategy performs switching between local tracking and global detection according to the previous tracking result, named local-global switching strategy. A merit of this strategy is that the temporal context including historical appearances and locations of the target can be exploited for local tracking. However, whether to switch to global detection is totally decided by the local tracking result. It increases the risk that the algorithm misidentifies a distractor as the target instead of activating the re-detector when the real target is out of the view of the local tracker.

In contrast to the local-global switching strategy, another type of long-term trackers~\cite{GlobalTrack,DMTrack} adopt a global tracking strategy which performs global re-detection to locate the target in the entire image on every frame. For example, GlobalTrack~\cite{GlobalTrack} performs global tracking by one-shot detection, totally ignoring the temporal context such as historical appearances and locations of the target. As a result, GlobalTrack is vulnerable to the appearance variations of the target and the background distractors. To alleviate the issue, DMTrack~\cite{DMTrack} introduces a Re-ID embedding into the global re-detection framework to associate the detections across adjacent frames to utilize the previous re-detection results. However, this Re-ID embedding is learned using only pedestrian datasets~\cite{ReIDinWild,PersonSearch,MOT16,PedestrianBenchmark}, limiting the generalization ability for tracking an arbitrary target object.

In this paper, we propose to perform global tracking via ensemble of local trackers, which combines the merits of both above strategies: tracking the target in a global view while exploiting the temporal context. Specifically, our algorithm deploys an ensemble of local trackers on different reference positions over the entire image, and every local tracker searches for the target within a local region around the reference position. With reasonable reference positions and search range, the search regions of all local trackers can cover the entire image, and then these local trackers collaborate to achieve global tracking.

The collaboration mechanisms in our design are: 1) When a local tracker successfully locates the target (called \emph{activated}), it will move following the target to attempt to keep tracking it in the subsequent frames. Generally, the target moving smoothly can be tracked by a local tracker consecutively. 2) When the activated local tracker accidentally loses the target due to suddenly discontinuous moving, another local tracker close to the target can readily take over the tracking to locate the target, i.e., being activated, to avoid tracking failure. The local tracker that loses the target will move back to where it starts moving. During the period when a local tracker steadily tracks the target, our algorithm exploits the temporal context to improve the local tracking robustness, which further extends the duration that this local tracker successfully tracks the target.

Specifically, we design a deformable attention-based local tracker to search for the target within a dynamic local region. Thus we can simulate the routine local tracking mechanism in the global view by moving the dynamic local search region following the target. Based on the local tracker, we propose a temporal context transferring scheme to exploit the historical appearances and locations of the target for local tracking. Figure~\ref{Fig:Introduction} illustrates the tracking process 
via ensemble of local trackers.
To conclude, we make the following contributions: 
\begin{itemize}
    \vspace{-1.75mm}
    \item We propose a global tracking algorithm via ensemble of local trackers, which can track the target in a global view while exploiting the temporal context.
    \vspace{-1.75mm}
    \item We design a deformable attention-based local tracker to simulate the local tracking mechanism in the global view and a temporal context transferring scheme based on the local tracker to exploit the temporal context.
    \vspace{-1.75mm}
    \item We achieve favorable performance against state-of-the-art methods on six diverse datasets, demonstrating the effectiveness of our algorithm.
\end{itemize}

\section{Related Work}
\noindent\textbf{Local-global Switching Strategy Trackers.}
Many algorithms~\cite{TLD,LCT,Fucolot} address the long-term tracking task through the local-global switching strategy. They equip the local tracker with a global re-detector to re-detect the target after local tracking fails, i.e., performing switching between local tracking and global re-detection based on the previous tracking result. TLD~\cite{TLD} is an early method using the strategy. It uses the optical flow for local tracking and an ensemble of weak classifiers for global re-detection. Recently, several methods~\cite{MBMD,LTMU,SPLT} introduce advanced deep local trackers and global re-detectors into this framework. Besides, some methods~\cite{DaSiamRPN,KeepTrack} that choose to enlarge the search region instead of global re-detection when local tracking fails can be seen as variants of this strategy.

A key issue for this strategy is how to decide whether to switch to global re-detection (or switch to a larger search region). Several methods~\cite{LCT,Fucolot,DaSiamRPN,KeepTrack} directly make the switching decision based on the response map predicted by the local tracker, while other approaches~\cite{MBMD,LTMU,SPLT} adopt an additional learnable verifier to manage the local-global switching. However, whether to switch from local tracking to global re-detection is still totally decided by the local tracking prediction. That is, the information out of the local search region is ignored for making the switch decision. It increases the risk that the algorithm misidentifies a distractor as the target instead of activating the global re-detector when the real target is out of the local search region.

\begin{figure*}[t]
\centering
    \includegraphics[width=0.93\textwidth]{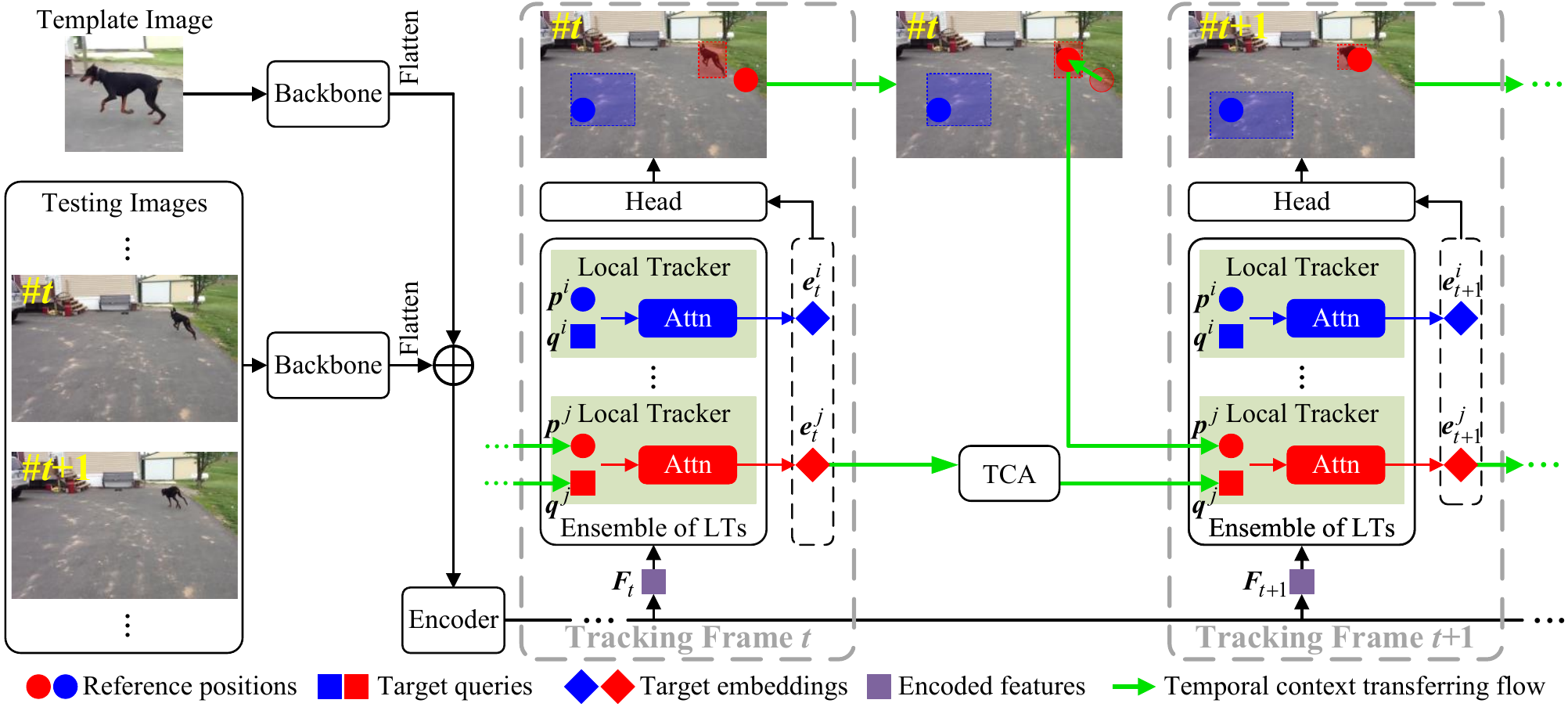}
    \vspace{-1.5mm}
    \caption{\textbf{The proposed global tracking framework via ensemble of local trackers.} It first extracts the backbone features for the template image and the testing images, and then adopts an encoder to enhance the target information in the features of the testing frames, generating the encoded feature for every testing image. An ensemble of local trackers (LTs) implemented with a decoder is constructed on the encoded feature to perform ensemble local tracking. $\oplus$ denotes the concatenation operation.}
\label{Fig:Framework}
\vspace{-4mm}
\end{figure*}

\vspace{1mm}
\noindent\textbf{Global Tracking Strategy Trackers.}
Several methods~\cite{BGDT,GlobalTrack,DMTrack,SiamRCNN} adopt a global tracking strategy by global re-detection on every frame for long-term tracking. Among these methods, GlobalTrack~\cite{GlobalTrack} performs tracking by global one-shot detection without considering the temporal context, which is sensitive to the target appearance variations. Siam R-CNN~\cite{SiamRCNN} designs a sophisticated global re-detector and associates the detections with a hand-crafted score for dynamic programming, but runs slowly due to the heavy computing burdens. Inspired by the MOT algorithms, DMTrack~\cite{DMTrack} introduces a Re-ID branch into the global re-detection framework to associate detections across frames. However, the Re-ID branch is trained using only human datasets~\cite{ReIDinWild,PersonSearch,MOT16,PedestrianBenchmark}, inevitably damaging the generalization ability. Unlike these global trackers that ignore the temporal context or exploit the temporal context after re-detection by association, our method performs global tracking via ensemble of local trackers, directly exploiting the temporal context for predicting the candidate.

\vspace{1mm}
\noindent\textbf{Transformer Tracking.} Recently, several transformer-based trackers~\cite{TransTrack,STARK,TransMT} have been proposed. Our idea is also implemented in an encoder-decoder structure like these works, but different from these methods that are designed as local trackers, our method aims to deploy an ensemble of local trackers in the encoder-decoder structure to achieve global tracking. Besides, some methods~\cite{Transformer,TransTrack,MOTR} adopt the encoder-decoder framework for MOT. Whilst our method and these MOT methods both use multiple queries, the major difference is: every query in MOT is in charge of detecting and tracking different objects, while in our method all queries work collaboratively for 
tracking the only target.
\vspace{-2mm}

\section{Method}
\vspace{-1mm}
In this section, we present our global tracking algorithm via ensemble of local trackers, which performs global tracking on every frame while effectively exploiting the temporal context. To track the target in a global view, our method deploys an ensemble of local trackers spreading over the entire image. Every local tracker searches for the target within different local regions. With reasonable distribution, the local search regions can together cover the entire image. Thus, these local trackers can perform global tracking via collaboration. Besides, the temporal context can be exploited when a local tracker continuously tracks the target.

\subsection{Global Tracking Framework}
\vspace{-1mm}
\label{Sec:Framework}
Figure~\ref{Fig:Framework} illustrates the global tracking framework of our method. In the following, we briefly introduce the tracking framework from two aspects: 1) feature extracting and encoding; 2) tracking by ensemble of local trackers.

\vspace{1mm}
\noindent\textbf{Feature Extracting and Encoding.}
Taken as input the template image $\bm I_{0}$ (cropped from the initial image) and a sequence of testing images $\{\bm I_{t}\}_{t=1}^{T}$, our method first extracts their backbone features $\bm F^{b}_{0} \in \mathbb{R}^{H_{z} \times W_{z} \times C}$ and $\{\bm F^{b}_{t} \in \mathbb{R}^{H_{s} \times W_{s} \times C}\}_{t=1}^{T}$ with a backbone network. For efficiency, we then use a linear layer to reduce the channel number of the backbone feature from $C$ to $c$.

To encode the information of the target to be tracked into the testing images, we use a transformer encoder stacking multi-head self-attention modules to perform fusion between the template image and the testing image, which has been proven to be effective by~\cite{STARK, TransTrack}. Specifically, the features of $\bm I_{0}$ and $\bm I_{t}$ are concatenated after a flatten operation and fed into the encoder. Then, the feature pixels corresponding to $\bm I_{t}$ are retrieved from the output of the encoder and reshaped to a 3-D tensor denoted by $\bm F_{t} \in \mathbb{R}^{H_{s} \times W_{s} \times c}$, in which the target information is enhanced.

\vspace{1mm}
\noindent\textbf{Tracking by Ensemble of Local Trackers.}
To achieve global tracking, we deploy an ensemble of local trackers to search for the target in different local regions in parallel. During tracking, if a local tracker is activated, i.e., locating the target, it will move following the target to keep tracking it in the subsequent frames. When the target moves smoothly, the activated local tracker can keep tracking the target in consecutive frames, forming an activated local tracker flow. Along with such flow, we can readily transfer and exploit the temporal context from multiple historical frames to perform local tracking in our framework. On the other hand, when the target moves discontinuously due to occlusion or disappearance, although the activated local tracker may lose the target, another local tracker close to the target can take over the tracking to locate the target.

In particular, we propose a deformable attention-based local tracker, which maintains a reference position and a target query to perform local tracking on top of the encoded feature $\bm F_{t}$. Every local tracker outputs a target embedding used for further predicting a candidate by a prediction head. Based on the local tracker, we design a temporal context transferring scheme, which transfers the target information using the reference position and target query as carriers along with the activated local tracker flow. 

Denoting the reference position and the target query of the $i$-th local tracker by $\bm p^{i}$ and $\bm q^{i}$, respectively, the tracking process with $N$ local trackers on $\bm I_{t}$ can be formulated as:
\begin{equation}
\setlength{\abovedisplayskip}{4pt}
\setlength{\belowdisplayskip}{4pt}
\label{Eq:tracking}
\begin{split}
    & \{\bm e^{i}_{t}\}_{i=1}^{N} = \Phi_{\rm LT}(\{\bm q^{i}\}_{i=1}^{N}, \{\bm p^{i}\}_{i=1}^{N}\},\bm F_{t}),\\
    & \{y^{i}_{t}\}_{i=1}^{N}=\Phi_{\rm Head}(\{\bm e^{i}_{t}\}_{i=1}^{N}).
\end{split}
\end{equation}
Herein $\bm e^{i}_{t}$ denotes the output target embedding of the $i$-th local tracker. $y^{i}_{t}\!=\!\{s^{i}_{t},\bm b^{i}_{t}\}$ is the candidate predicted based on $\bm e^{i}_{t}$, where $s^{i}_{t}$ is the foreground-background classification score and $\bm b^{i}_{t}$ is the bounding box. $\Phi_{\rm LT}$ and $\Phi_{\rm Head}$ denote the parallel local trackers and the head, respectively. Assuming the $j$-th local tracker is activated on $\bm I_{t}$, the temporal context transferring scheme can be formulated as:
\begin{equation}
\setlength{\abovedisplayskip}{4pt}
\setlength{\belowdisplayskip}{4pt}
\label{Eq:transferring}
    \bm p^{j} \gets \mathcal T_{\rm rp}(y^{j}_{t}), \bm q^{j} \gets \mathcal T_{\rm tq}(\bm e^{j}_{t}),
\end{equation}
where $\mathcal T_{\rm rp}$ and $\mathcal T_{\rm tq}$ denote the temporal context transferring via the reference position and the target query, respectively. The updated $\bm p^{j}$ and $\bm q^{j}$ are then used for tracking on $\bm I_{t+1}$.

\subsection{Deformable Attention-based Local Trackers}
\vspace{-1mm}
\label{Sec:Local_Tracker}
To perform local tracking within the global view, our local tracker should be able to search for the target in a dynamic local region on the encoded feature $\bm F_{t}$ of the testing image $\bm I_{t}$. So that we can move the activated local tracker following the target by changing its search region according to the previous tracking result. To this end, we opt for the Deformable Attention~\cite{Deformable_DETR} to implement our local tracker due to its ability of adaptive and sparse sampling around a reference location. Thus, we can move the local tracker by setting a new reference position for it.

Every local tracker maintains a target query and a reference position. The target query models the potential target information in appearance and location, while the reference position determines the search region of the local tracker coarsely. In particular, the default target query of a local tracker is an offline learned embedding, and the default reference position is predicted from the target query via a linear layer and sigmoid function. To perform local tracking, a local tracker computes the attention between its target query and the feature pixels sampled from $\bm F_{t}$ around its reference position, in which the sample positions are produced from its target query by predicting the coordinate offsets to its reference position via a linear layer. Through the attention operation, every local tracker outputs a target embedding that models the appearance and location information of a candidate in the corresponding local search region. Technically, all local trackers are implemented with a decoder, and the parallel local tracking process $\Phi_{\rm LT}$ is defined as:
\begin{equation}
\setlength{\abovedisplayskip}{4pt}
\setlength{\belowdisplayskip}{4pt}
\label{Eq:deform_attn}
\bm \Phi_{\rm LT}(\{\bm q^{i}\}_{i=1}^{N}, \{\bm p^{i}\}_{i=1}^{N}\},\bm F_{t})={A_{dc}}({A_{ms}}(\bm q),\bm p,\bm F_{t}),
\end{equation}
where $A_{dc}$ and $A_{ms}$ denote the deformable cross-attention~\cite{Deformable_DETR} and the multi-head self-attention~\cite{Transformer} functions, respectively. $\bm q=\bm q^{1}\oplus \cdots \oplus \bm q^{N}$, $\bm p=\bm p^{1}\oplus \cdots \oplus \bm p^{N}$, and $\oplus$ denotes the concatenation operation.

It is worth noting that the self-attention operation that models the interaction between all target queries promotes the learned reference positions of all local trackers to spread over the image reasonably during training, which is crucial for learning an effective arrangement of local trackers.

\subsection{Temporal Context Transferring}
\vspace{-1mm}
\label{Sec:Temporal_Context}
To exploit the temporal context in our global tracking framework, we use the reference position and the target query as the carriers to transfer the temporal context along with the activated local tracker flow. Next, we detail this process assuming that the $j$-th local tracker keeps activated.

The predicted target location in the previous frame is a direct clue denoting where to search for the target locally in the subsequent frame. We therefore directly use it as the new reference position of the activated local tracker for tracking the subsequent frame. Thus, $\mathcal T_{\rm rp}$ is defined as:
\begin{equation}
\setlength{\abovedisplayskip}{4pt}
\setlength{\belowdisplayskip}{4pt}
\label{Eq:transfer_rp}
\mathcal T_{\rm rp}(y^{j}_{t})=\bm c^{j}_{t},
\end{equation}
where $\bm c^{j}_{t}$ is the center of the predicted target bounding box $\bm b^{j}_{t}$ in the previous testing image $\bm I_{t}$. In this way, the activated local tracker will move following the target until the target is located by another local tracker or disappears. After losing the target, the reference position of this local tracker will be reset to its default value. Namely, this local tracker will move back to where it starts moving.

In addition, the target embedding predicted by an activated local tracker contains the new target information in appearance and location, and thus it can be naturally used to generate a new target query for the activated local tracker to track in the subsequent frame. To this end, we propose a Temporal Context Aggregation (TCA) model to aggregates the temporal context modeled in the recently predicted target embeddings to generate a new target query. We refer to such a new target query as the \emph{online target query} for clarity.

\begin{figure}[t]
\centering
    \includegraphics[width=0.975\columnwidth]{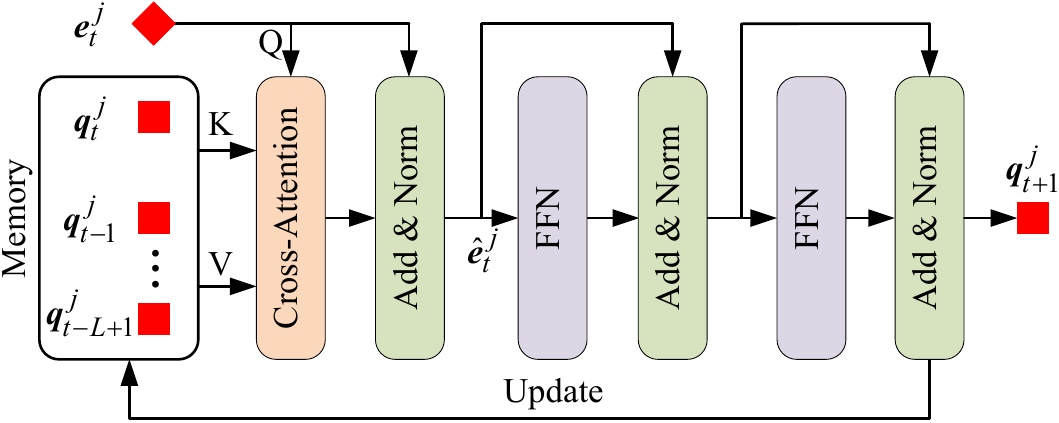}
    \vspace{-1.0mm}
    \caption{\textbf{Architecture of the proposed temporal context aggregation model.} It performs the interaction between the target embedding and the historical online target queries to aggregate the temporal context information modeled by these queries.}
\label{Fig:TC}
\vspace{-4mm}
\end{figure}

As shown in Figure~\ref{Fig:TC}, our TCA model maintains a memory of the online target queries generated in the recent $L$ frames $\bm q^{j}_{mem}=\bm q_{t-L+1}^{j}\oplus \cdots \oplus \bm q_{t}^{j}$. Such a memory models the historical appearance and trajectory information of the target in the recent frames. To aggregate this historical information, the TCA model computes the cross-attention between the target embedding $\bm e^{j}_{t}$ (serving as the query) and the memory $\bm q^{j}_{mem}$ (serving as the key and value) with a multi-head cross-attention layer. Then, the output of cross-attention and the target embedding $\bm e^{j}_{t}$ are added and normalized to generate an aggregated target embedding $\hat{\bm e}^{j}_{t}$. Finally, two feed-forward networks (FFNs) with a skip connection followed by a normalization layer are used to adjust $\hat{\bm e}^{j}_{t}$ to generate the online target query $\bm q^{j}_{t+1}$ used for tracking on $\bm I_{t+1}$. Formally, $\mathcal T_{\rm tq}$ can be defined as:
\begin{equation}
\setlength{\abovedisplayskip}{4pt}
\setlength{\belowdisplayskip}{4pt}
\label{Eq:temporal_context}
\mathcal T_{\rm tq}(\bm e^{j}_{t})=\phi_{adj}(\phi_{norm}(A_{mc}(\bm e^{j}_{t}, \bm q^{j}_{mem})+\bm e^{j}_{t})),
\end{equation}
where $A_{mc}$, $\phi_{norm}$ and $\phi_{adj}$ denote the cross-attention, normalization, and adjustment operations, respectively. We also use $\bm q^{j}_{t+1}$ to update the memory and pop out the oldest query. For a newly activated local tracker flow, the memory of the TCA model is empty. In this case, we direct feed the target embedding into the last two FFNs in Figure~\ref{Fig:TC} to generate an online target query, omitting the other layers.

\subsection{Supervised Learning of Local Trackers}
\vspace{-1mm}
To learn the temporal context modeling ability, we train the proposed model with sequence samples. In a sequence sample, the target patch cropped from the first frame is used as the template image, and the subsequent frames are used as the testing images. During training, we perform the forward propagation on the testing images following the pipeline shown in Figure~\ref{Fig:Framework}.

To calculate the loss on the $t$-th testing image $\bm I_{t}$, we first use the Hungarian algorithm~\cite{Hungarian} to match the ground truth $\hat y_{t} = \{{\hat {\bm b}}_{t}\}$ with a predicted candidate $y^{i}_{t}=\{s^{i}_{t},\bm b^{i}_{t}\}$. Herein ${\hat {\bm b}}_{t}$ denotes the ground truth bounding box.
Based on the Hungarian loss in DETR~\cite{DETR}, we further consider the $\ell_{1}$-norm of the distance between the ground truth box center ${\hat {\bm p}}_{t}$ and the reference position $\bm p^{i}$ of the local tracker as a regularization. This regularization encourages the candidate predicted by the local tracker close to the target to be matched with the ground truth, benefiting the learning of our model in two aspects. First, it increases the probability that a local tracker keeps activated in continuous testing images, which is necessary for learning the temporal context modeling. Second, it benefits the local tracker to learn a reasonable deformable attention range. Otherwise, the local trackers tend to over-extend the deformable attention range in training, which would cause a local tracker to perceive too much background information and thus become vulnerable. Our Hungarian loss $\mathcal {L}_{H}(y^{i}_{t},{\hat y}_{t})$ is defined as:
\begin{equation}
\setlength{\abovedisplayskip}{4pt}
\setlength{\belowdisplayskip}{4pt}
\label{Eq:hungarian_loss}
\resizebox{0.905\hsize}{!}{$
\begin{aligned}
    &\mathcal {L}_{H}(y^{i}_{t},{\hat y}_{t})\!=\!\lambda_{cls}\mathcal L_{cls}(s^{i}_{t})\!+\!\mathcal {L}_{box}(\bm b^{i}_{t},{\hat {\bm b}}_{t})\!+\!\lambda_{r}\mathcal {L}_{\ell_{1}}(\bm p^{i},{\hat {\bm p}}_{t}),\\
    &\mathcal {L}_{box}(\bm b^{i}_{t},{\hat {\bm b}}_{t})\!=\!\lambda_{\ell_{1}}\mathcal {L}_{1}(\bm b^{i}_{t},{\hat {\bm b}}_{t})\!+\!\lambda_{iou}\mathcal {L}_{iou}(\bm b^{i}_{t},{\hat {\bm b}}_{t}).
\end{aligned}
$}
\end{equation}
Here, $\mathcal L_{cls}$, $\mathcal L_{\ell_{1}}$, and $\mathcal L_{iou}$ refer to the focal loss~\cite{FocalLoss}, $\ell_{1}$ loss, and generalized IoU loss~\cite{GIoU}, respectively. $\lambda_{cls}$, $\lambda_{r}$, $\lambda_{\ell_{1}}$, and $\lambda_{iou}$ are the balance weights. Note that we compute bipartite matching in every testing image instead of directly propagating the assignment in the first testing image to the following ones, as it also leads to the over-extension of the deformable attention range.

Denoting the index of the matched candidate in the $t$-th testing image as $\pi_{t}$, the training loss $\mathcal {L}$ for a sequence sample with $T$ testing images can be formulated as:
\begin{equation}
\setlength{\abovedisplayskip}{4pt}
\setlength{\belowdisplayskip}{4pt}
\label{Eq:training_loss}
\mathcal {L}\!=\!\frac{1}{T}\sum_{t=1}^{T}\sum_{i=1}^{N}\lambda_{cls}\mathcal L_{cls}(s^{i}_{t})\!+\!\mathds{1}_{\{i=\pi_{t}\}}\mathcal {L}_{box}(\bm b_{t},{\hat {\bm b}}^{i}_{t}),
\end{equation}
where $\mathds{1}_{\{i=\pi_{t}\}}$ equals $1$ if $i\!=\!\pi_{t}$ and $0$ otherwise.

\section{Experiments}
\subsection{Implementation Details}
\vspace{-1mm}
We adopt ResNet-50~\cite{ResNet} pre-trained on ImageNet~\cite{Imagenet} as our backbone, and the output of \emph{conv}-4 is used as the backbone feature. We use the head model in DETR~\cite{DETR} to predict the candidate box and the corresponding classification score. Besides, we compute the cosine similarity between the target template and the candidate in the feature space, and then obtain the candidate confidence by multiplying the classification score and the cosine similarity. Similar to~\cite{DMTrack}, We adopt a Hungarian algorithm~\cite{Hungarian} to select the final prediction from the candidates, which takes the confidence and the position of the candidate into account. Whether the target is present is determined by comparing the confidence score with a threshold $\theta$. The template image is cropped from the initial frame centered on the ground truth target and then resized to $128 \times 128$, whose area is $2^2$ times that of the target. The testing images are resized to $640 \times 480$. The local tracker number $N$ and the memory length $L$  are set to ten and five by default, respectively.

During training, the length of the sequence sample (including one template image and one or more testing images) gradually increases from 2 to 6. We use the training splits of COCO~\cite{COCO}, LaSOT~\cite{LaSOT}, TrackingNet~\cite{TrackingNet}, and GOT-10k~\cite{Got-10k} to train our model, and COCO is only used when the length of the sequence sample is 2.
$\lambda_{cls}$, $\lambda_{r}$, $\lambda_{\ell_{1}}$, and $\lambda_{iou}$ are set to 1.0, 5.0, 5.0, 2.0, respectively.
Source codes will be available at https://github.com/ZikunZhou/GTELT.

\begin{table}[t]
\begin{center}
\caption{\textbf{Precision (Pre.), normalized precision (nPre.), and AUC for four variants of our method on LaSOT.}}
\label{Tab:Abla}
\vspace{-2.5mm}
\scriptsize
\renewcommand\arraystretch{1.}
\resizebox{0.8\linewidth}{!}{
\begin{tabular}{l|ccccc}
\toprule
Variants  & Our model & OSDet & EGT & SGT \\
\midrule
Pre. & 0.732 & 0.707 & 0.690 & 0.670 \\
nPre. & 0.759 & 0.732 & 0.717 & 0.705 \\
AUC & 0.677 & 0.653 & 0.623 & 0.619\\ 
\bottomrule
\end{tabular}}
\end{center}
\vspace{-8mm}
\end{table}
\subsection{Ablation Study}
\vspace{-1mm}
We first conduct experiments to investigate the effectiveness of each proposed technique in our model. To this end, we perform ablation studies on four variants of our method:

\noindent1) \textbf{Our model}, our intact model that performs global tracking via ensemble of local trackers.

\noindent2) \textbf{OSDet}, which removes the temporal context transferring scheme from our model. Thus, our model degenerates into a global \textbf{O}ne-\textbf{S}hot \textbf{Det}ector in the transformer framework, using multiple offline learned queries to perform global re-detection without exploiting the temporal context.

\noindent3) \textbf{EGT}, replacing the deformable attention module in our model with the multi-head attention module, which computes attention globally and densely. Thus, the ensemble of local trackers in our model becomes an \textbf{E}nsemble of \textbf{G}lobal \textbf{T}rackers. Due to these global trackers do not involve reference positions, we perform temporal context transferring only using the target queries as the carriers.

\noindent4) \textbf{SGT}, reducing the number of target queries in EGT to one, i.e., adopting a \textbf{S}ingle multi-head attention-based \textbf{G}lobal \textbf{T}racker to perform tracking.

Table~\ref{Tab:Abla} presents the experimental results of these variants on the testing set of LaSOT~\cite{LaSOT}.

\noindent \textbf{Effect of the Temporal Context Transferring Scheme.}
The performance gap between our model and OSDet clearly demonstrates the effectiveness of the proposed temporal context transferring scheme for exploiting the historical appearances and locations of the target in global tracking.

\noindent \textbf{Effect of the Deformable Attention-based Local Tracker.} Compared with our method, the performance of EGT and SGT decreases by 5.4\% and 5.8\% in AUC, respectively. We attribute the performance drops to two reasons: 1) EGT and SGT cannot use the previous target location as a direct clue for locating the target in the subsequent frame; 2) Every query in EGT and SGT interacts with all feature pixels in $\bm F_{t}$, and thus the background information from the entire image inevitably overwhelms the target information. In addition, EGT using multiple queries performs marginally better than SGT using a single query, which is different from the result reported by STARK~\cite{STARK}. In STARK, using multiple queries leads to a performance drop compared with using a single query. We guess the reason for this difference is that more than one query is needed for locating the target when the search region becomes the entire image.

\noindent\textbf{Effect of the local tracker number $N$ and the memory length $L$.} We also conduct experiments to investigate the effect of $N$ and $L$. Table~\ref{Tab:Ablation_NL} reports the results of our model by varying $N$ or $L$ on LaSOT. We can observe that both sparse ($N=5$) and dense ($N=20$) local trackers damage the performance. We guess the reason is that sparse local trackers can hardly cover the full image while dense local trackers can hardly receive sufficient training since in each frame only the activated one receives the feedback from the supervision. Besides, tracking performance improves and saturates at about $L=7$ along with increasing $L$.

\begin{table}[t]
\setlength\tabcolsep{4.0pt}
\begin{center}
\caption{\textbf{Ablation studies on the local tracker number ($N$) and the memory length ($L$) on LaSOT.}}
\label{Tab:Ablation_NL}
\vspace{-2.5mm}
\scriptsize
\renewcommand\arraystretch{1.0}
\resizebox{1.0\linewidth}{!}{
\begin{tabular}{l|ccc|ccccc}
\toprule
           & \multicolumn{3}{c|}{$L$=5} & \multicolumn{5}{c}{$N$=10} \\
           & $N$=5 & $N$=10 & $N$=20 & $L$=1 & $L$=3 & $L$=5 & $L$=7 & $L$=9 \\
\midrule
Pre.~  & 0.712 & 0.732 & 0.705 & 0.725 & 0.726 & 0.732 & 0.731 & 0.730\\
nPre.~ & 0.743 & 0.759 & 0.734 & 0.755 & 0.755 & 0.759 & 0.759 & 0.757\\ 
AUC~   & 0.670 & 0.677 & 0.660 & 0.672 & 0.675 & 0.677 & 0.677 & 0.675\\
\bottomrule
\end{tabular}}
\vspace{-8mm}
\end{center}
\end{table}

\vspace{-1mm}
\subsection{Comparison with State-of-the-art Trackers}
\vspace{-1mm}
We compare our algorithm with state-of-the-art trackers on six datasets including TLP~\cite{TLP}, OxUvA~\cite{OxUvA}, VOT2020-LT~\cite{VOT2020}, LaSOT~\cite{LaSOT}, LaSOTExtSub~\cite{LaSOTExtSub}, and TrackingNet~\cite{TrackingNet}. The trackers involved in the comparison include three global trackers (DMTrack~\cite{DMTrack}, Siam R-CNN~\cite{SiamRCNN}, and GlobalTrack~\cite{GlobalTrack}), eight local-global switching strategy trackers (KeepTrack~\cite{KeepTrack}, KeepTrackFast~\cite{KeepTrack}, LT\_DSE~\cite{VOT2020}, CLGS~\cite{VOT2020}, LTMU~\cite{LTMU}, SPLT~\cite{SPLT}, MBMD~\cite{MBMD}, and TLD~\cite{TLD}), and nine local trackers (STARK~\cite{STARK}, TransT~\cite{TransTrack}, TrDiMP~\cite{TransMT}, PrDiMP~\cite{PrDiMP}, AlphaRefine~\cite{AlphaRefine}, SuperDiMP~\cite{SuperDiMP}, MDNet~\cite{MDNet}, STMTrack~\cite{STMTrack}, and SiamFC~\cite{SiamFC}). We discuss the experimental results per dataset and the running speed below.

\vspace{1mm}
\noindent\textbf{TLP.} TLP~\cite{TLP} includes 50 long sequences with an average sequence length of about 13,500 frames. Table~\ref{Tab:TLP} reports the AUC and precision scores on TLP. Compared with the local-global switching strategy tracker LTMU, our method improves the performance by 1.2\% in AUC and 0.9\% in precision. Besides, our method outperforms the other two global trackers, DMTrack and GlobalTrack, by a large margin (2.9\%/6.9\% in AUC and 2.0\%/4.4\% in precision, respectively), demonstrating the effectiveness of our method.

\begin{figure}[t]
\centering
\includegraphics[width=1.60in]{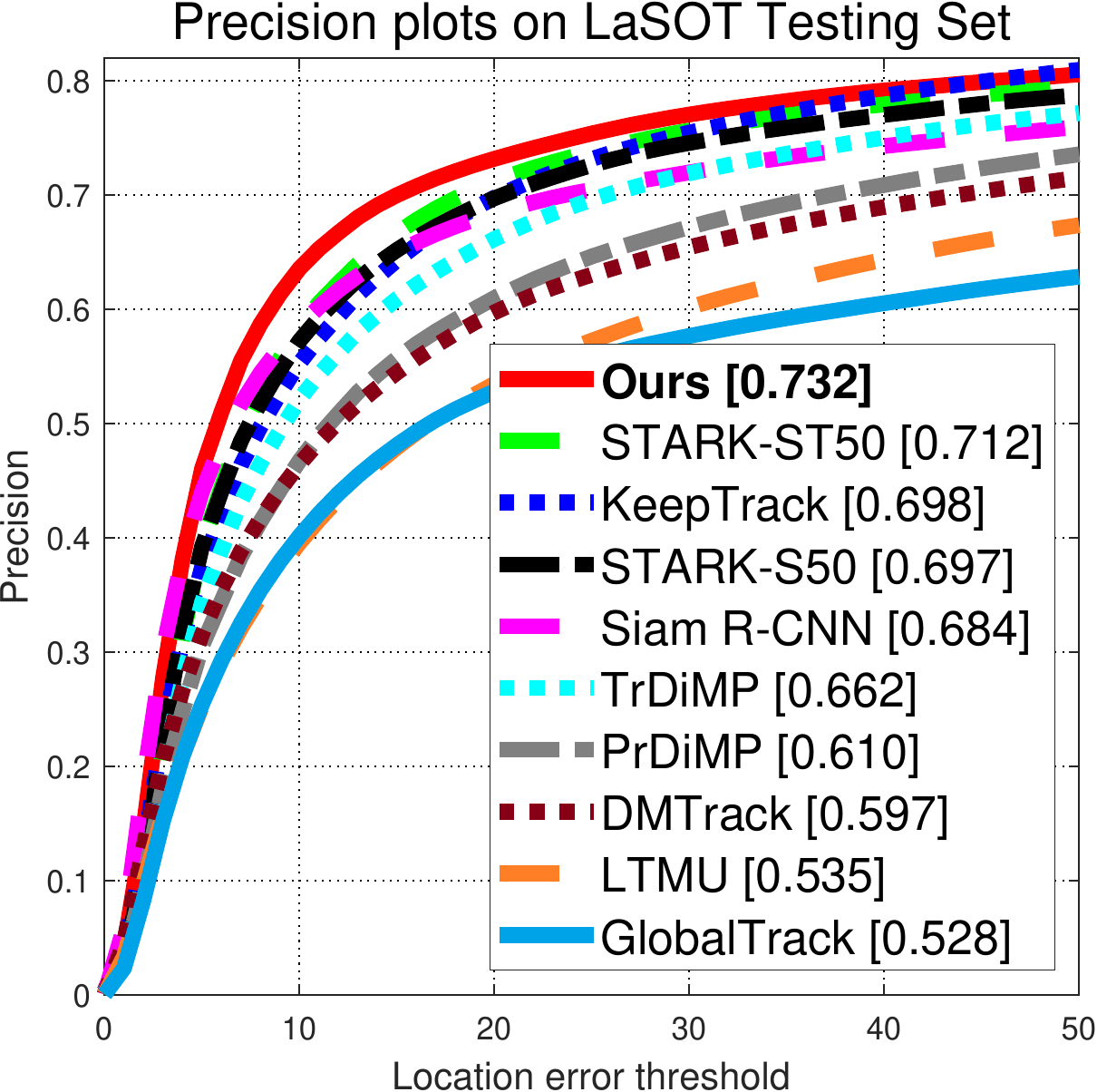}
\includegraphics[width=1.60in]{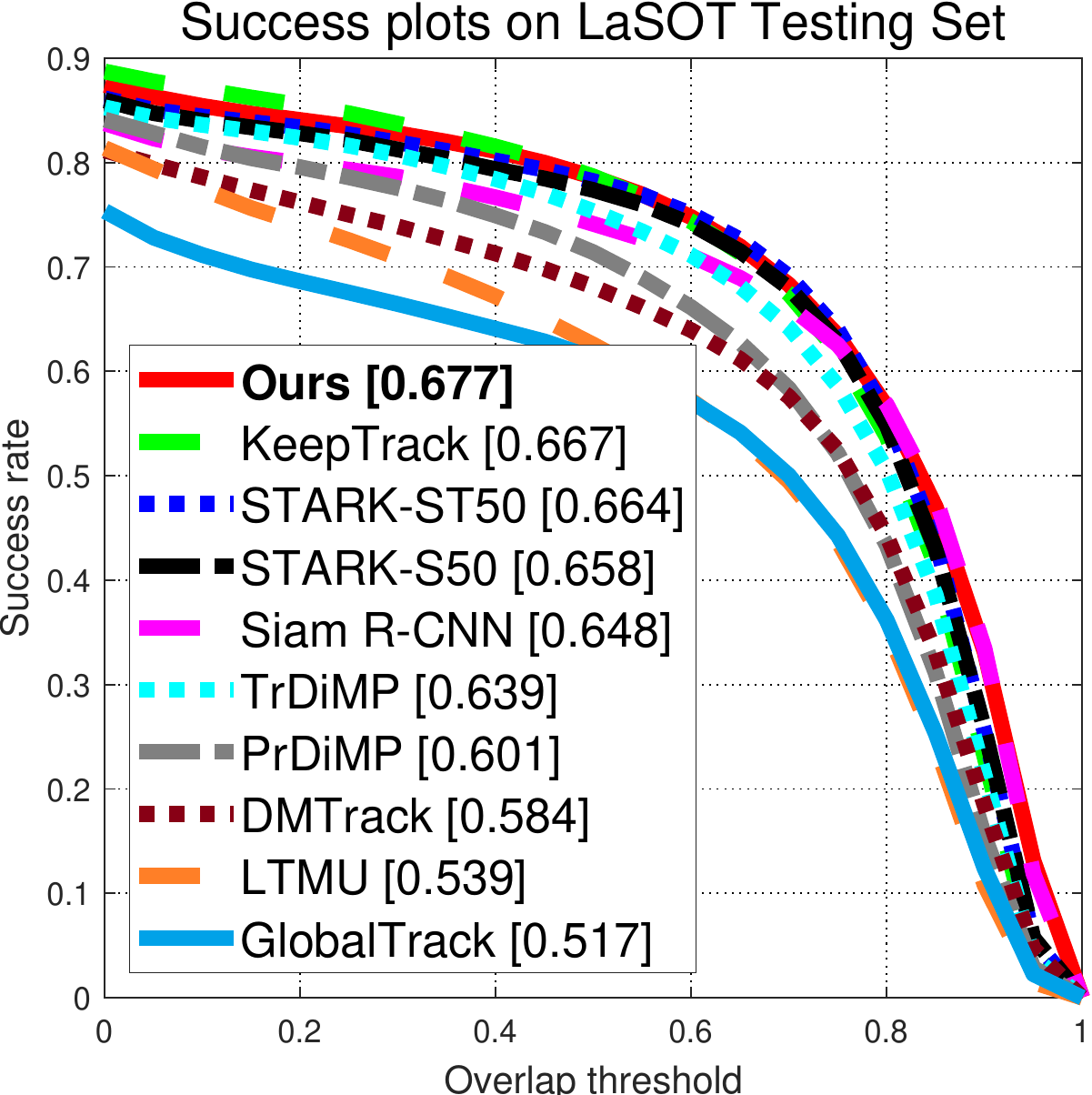}
\vspace{-2.mm}
\caption{\textbf{Precision and success plots of different algorithms on the LaSOT testing set.}}
\vspace{-3mm}
\label{Fig:LaSOT}
\end{figure}

\vspace{1mm}
\noindent\textbf{OxUvA.} OxUvA~\cite{OxUvA} contains 166 long testing sequences. Besides the target bounding box, the OxUvA benchmark also requires trackers to predict whether the target is present. It uses the true positive rate (TPR), the true negative rate (TNR), and the maximum geometric mean (MaxGM) of TPR and TNR as performance metrics. We set the threshold $\theta$ to 0.6 for the evaluation on OxUvA. Table~\ref{Tab:OxUvA} presents the experimental results. Compared with the global tracker Siam R-CNN, our method achieves substantial performance gains of 4.5\% in MaxGM. KeepTrack is a recently proposed local-global switching strategy tracker, which uses an association network to deal with the distractors and performs better than our approach.

\newcommand{\tabincell}[2]{\begin{tabular}{@{}#1@{}}#2\end{tabular}}
\begin{table}[t]
\setlength\tabcolsep{2.5pt}
\begin{center}
\caption{\textbf{AUC and precision of different trackers on TLP.} The best and second-best scores are marked by \textbf{bold} and \underline{underline}, respectively. From left to right, trackers are divided into local trackers, local-global switching strategy trackers, and global trackers.}
\label{Tab:TLP}
\vspace{-2mm}
\scriptsize
\renewcommand\arraystretch{1}
\resizebox{1.0\linewidth}{!}{
\begin{tabular}{l|cc|ccc|ccc}
\toprule
           &
           \tabincell{c}{SiamFC\\~\cite{SiamFC}} &
           \tabincell{c}{MDNet\\~\cite{MDNet}} & 
           \tabincell{c}{TLD\\~\cite{TLD}} &
           \tabincell{c}{SPLT\\~\cite{SPLT}} &  
           \tabincell{c}{LTMU\\~\cite{LTMU}} & \tabincell{c}{Global\\Track~\cite{GlobalTrack}} &
           \tabincell{c}{DMTrack\\~\cite{DMTrack}} &  \textbf{Ours} \\
\midrule
AUC~        & 0.235 & 0.370 & 0.154 & 0.416 & \underline{0.558} & 0.501 & 0.541 & \textbf{0.570} \\
Pre.~       & 0.284 & 0.384 & 0.167 & 0.403 & \underline{0.602} & 0.567 & 0.591 & \textbf{0.611} \\
\bottomrule
\end{tabular}}
\vspace{-5mm}
\end{center}
\end{table}

\begin{table}[t]
\setlength\tabcolsep{2.0pt}
\begin{center}
\caption{\textbf{TPR, TNR, and MaxGM of different trackers on the OxUvA testing set.} From left to right, the trackers are divided into local-global switching strategy trackers and global trackers.}
\label{Tab:OxUvA}
\vspace{-2mm}
\scriptsize
\renewcommand\arraystretch{1}
\resizebox{1.0\linewidth}{!}{
\begin{tabular}{l|cccc|cccc}
\toprule
           &\tabincell{c}{MBMD\\~\cite{MBMD}} &
           \tabincell{c}{SPLT\\~\cite{SPLT}} & 
           \tabincell{c}{LTMU\\~\cite{LTMU}} & 
           \tabincell{c}{Keep\\Track~\cite{KeepTrack}} &
           \tabincell{c}{Global\\Track~\cite{GlobalTrack}} &
           \tabincell{c}{DMTrack\\~\cite{DMTrack}} &   \tabincell{c}{Siam R-\\CNN~\cite{SiamRCNN}} &
           \textbf{Ours} \\
\midrule
TPR        & 0.609 & 0.498 & 0.749 & \textbf{0.806} & 0.574 & 0.686 & 0.701 &  \underline{0.764} \\
TNR        & 0.485 & \underline{0.776} & 0.754 & \textbf{0.812} & 0.633 & 0.694 & 0.745 & 0.772 \\
MaxGM      & 0.544 & 0.622 & 0.751 & \textbf{0.809} & 0.603 & 0.688 & 0.723 & \underline{0.768} \\
\bottomrule
\end{tabular}}
\vspace{-5mm}
\end{center}
\end{table}

\vspace{1mm}
\noindent\textbf{VOT2020-LT.} VOT2020-LT~\cite{VOT2020} is a popular long-term tracking benchmark containing 50 challenging sequences. The VOT2020-LT benchmark requires trackers to predict a target bounding box and a corresponding confidence score. Based on the two predictions, precision, recall, and F-score are used as performance metrics. Table~\ref{Tab:VOT2020-LT} reports the experimental results on VOT2020-LT. While KeepTrack achieves the best performance, our method performs on par with KeepTrackFast and sophisticated LT\_DSE (VOT2020-LT winner), demonstrating the potential of our method.

\begin{table}[t]
\setlength\tabcolsep{2.pt}
\begin{center}
\caption{\textbf{Precision, recall, and F-score of different trackers on the VOT2020-LT dataset.}}
\label{Tab:VOT2020-LT}
\vspace{-2mm}
\scriptsize
\renewcommand\arraystretch{0.95}
\resizebox{1.0\linewidth}{!}{
\begin{tabular}{l|c|cccc|cc}
\toprule
           & \tabincell{c}{Super\\DiMP~\cite{SuperDiMP}} 
           & \tabincell{c}{CLGS\\~\cite{VOT2020}} 
           & \tabincell{c}{KeepTrack\\Fast~\cite{KeepTrack}} 
           & \tabincell{c}{LT\_DSE\\~\cite{VOT2020}} 
           & \tabincell{c}{Keep\\Track~\cite{KeepTrack}} 
           & \tabincell{c}{DMTrack\\~\cite{DMTrack}} & \textbf{Ours} \\
\midrule
Precision~  & 0.676 & 0.739 & 0.706 & \underline{0.715} & \textbf{0.723} & 0.690 & 0.695\\
Recall~  & 0.663 & 0.619 & 0.680 & 0.677 & \textbf{0.697} & 0.662 & \underline{0.690}\\
F-score~     & 0.669 & 0.674 & 0.693 & \underline{0.695} & \textbf{0.709} & 0.687 & 0.693\\ 
\bottomrule
\end{tabular}}
\vspace{-7mm}
\end{center}
\end{table}

\vspace{1mm}
\noindent\textbf{LaSOT.} LaSOT~\cite{LaSOT} contains 280 sequences in the testing set with an average sequence length of about 2,500 frames. It uses success, precision, and normalized precision as metrics. Figure~\ref{Fig:LaSOT} shows the precision and success plots on LaSOT. Our method achieves the best AUC and precision scores. Compared with the other global trackers, Siam R-CNN and DMTrack, our method achieves performance gains of 2.9\%/9.3\% in AUC and 4.8\%/13.5\% in precision, respectively. Besides, our method performs favorably against KeepTrack by performance gains of 1.0\% in AUC and 3.4\% in precision. 

\begin{table}[t]
\setlength\tabcolsep{2.5pt}
\begin{center}
\caption{\textbf{AUC, precision, and normalized precision on LaSOTExtSub and the speed of different trackers.} All speeds are reported on the RTX 2080Ti GPU other than those denoted by $*$.}
\label{Tab:LaSOTExtSub}
\vspace{-2mm}
\scriptsize
\renewcommand\arraystretch{1}
\resizebox{1.0\linewidth}{!}{
\begin{tabular}{l|ccc|ccc|cc}
\toprule
           
           &\tabincell{c}{ATOM\\~\cite{ATOM}} & 
           \tabincell{c}{DiMP\\~\cite{DiMP}} &  
           \tabincell{c}{Super\\DiMP~\cite{SuperDiMP}} &
           \tabincell{c}{SPLT\\~\cite{SPLT}} &
           \tabincell{c}{LTMU\\~\cite{LTMU}} &
            \tabincell{c}{Keep\\Track~\cite{KeepTrack}} &
            \tabincell{c}{Global\\Track~\cite{GlobalTrack}} &\textbf{Ours} \\
\midrule
AUC~        & 0.376 & 0.392 & 0.433 & 0.272 & 0.414 & \textbf{0.482} & 0.356 & \underline{0.450} \\
Pre.~       & 0.430 & 0.451 & 0.505 & 0.297 & 0.473 & \textbf{0.564} & 0.411 & \underline{0.524} \\
nPre.~      & 0.459 & 0.476 & 0.524 & 0.339 & 0.499 & \textbf{0.581} & 0.436 & \underline{0.542} \\
\midrule
FPS         & \textbf{65}   & \underline{54} & 39 & $26^*$ & 13   & 18 & $6^*$ & 26  \\
\bottomrule
\end{tabular}}
\vspace{-5mm}
\end{center}
\end{table}

\vspace{1mm}
\noindent\textbf{LaSOTExtSub.} LaSOTExtSub~\cite{LaSOTExtSub} is an extension set of LaSOT containing 15 new classes with 10 sequences each. Many objects in LaSOTExtSub are small objects. Specifically, the percentage of small objects (whose area is smaller than $32^2$ after resizing the entire image to $640 \times 480$) in LaSOTExtSub is 53.8\%, while the number in LaSOT is 17.6\%. The experimental results are reported in Table~\ref{Tab:LaSOTExtSub}. KeepTrack performs better than our method. The reason is that KeepTrack resizes the search region whose area is $8^2$ times that of the target to a fixed size, and this operation will increase the target resolution when the target is small. By contrast, our method always resizes the entire image to $640 \times 480$, which is not conducive to tracking small objects. Nevertheless, our method still outperforms the other long-term trackers, such as LTMU and GlobalTrack.

\begin{table}[t]
\setlength\tabcolsep{1.5pt}
\begin{center}
\caption{\textbf{AUC, precision, and normalized precision of different trackers on TrackingNet.} From left to right, trackers are divided into local trackers and global trackers.}
\label{Tab:TrackingNet}
\vspace{-2mm}
\scriptsize
\renewcommand\arraystretch{1}
\resizebox{1.0\linewidth}{!}{
\begin{tabular}{l|ccccc|ccc}
\toprule
           & \tabincell{c}{TrDiMP\\~\cite{TransMT}} &
           \tabincell{c}{STM\\Track~\cite{STMTrack}} & 
           \tabincell{c}{Alpha\\Refine~\cite{AlphaRefine}} &
           \tabincell{c}{STARK-\\ST50~\cite{STARK}} & \tabincell{c}{TransT\\~\cite{TransTrack}} &  \tabincell{c}{Global\\Track~\cite{GlobalTrack}} & 
           \tabincell{c}{Siam R-\\CNN~\cite{SiamRCNN}} & \textbf{Ours} \\
\midrule
AUC~        & 0.784 & 0.803 & 0.805 & 0.813 & \underline{0.814} & 0.704 & 0.812 & \textbf{0.825} \\
Pre.~       & 0.731 & 0.767 & 0.783 &  --  & \underline{0.800} & 0.656 & \underline{0.800} & \textbf{0.816} \\
nPre.~      & 0.833 & 0.851 & 0.856 & \underline{0.861} & \textbf{0.867} & 0.754 & 0.854 & \textbf{0.867} \\ 
\bottomrule
\end{tabular}}
\vspace{-8mm}
\end{center}
\end{table}

\begin{figure*}[t]
\centering
\includegraphics[width=0.96\textwidth]{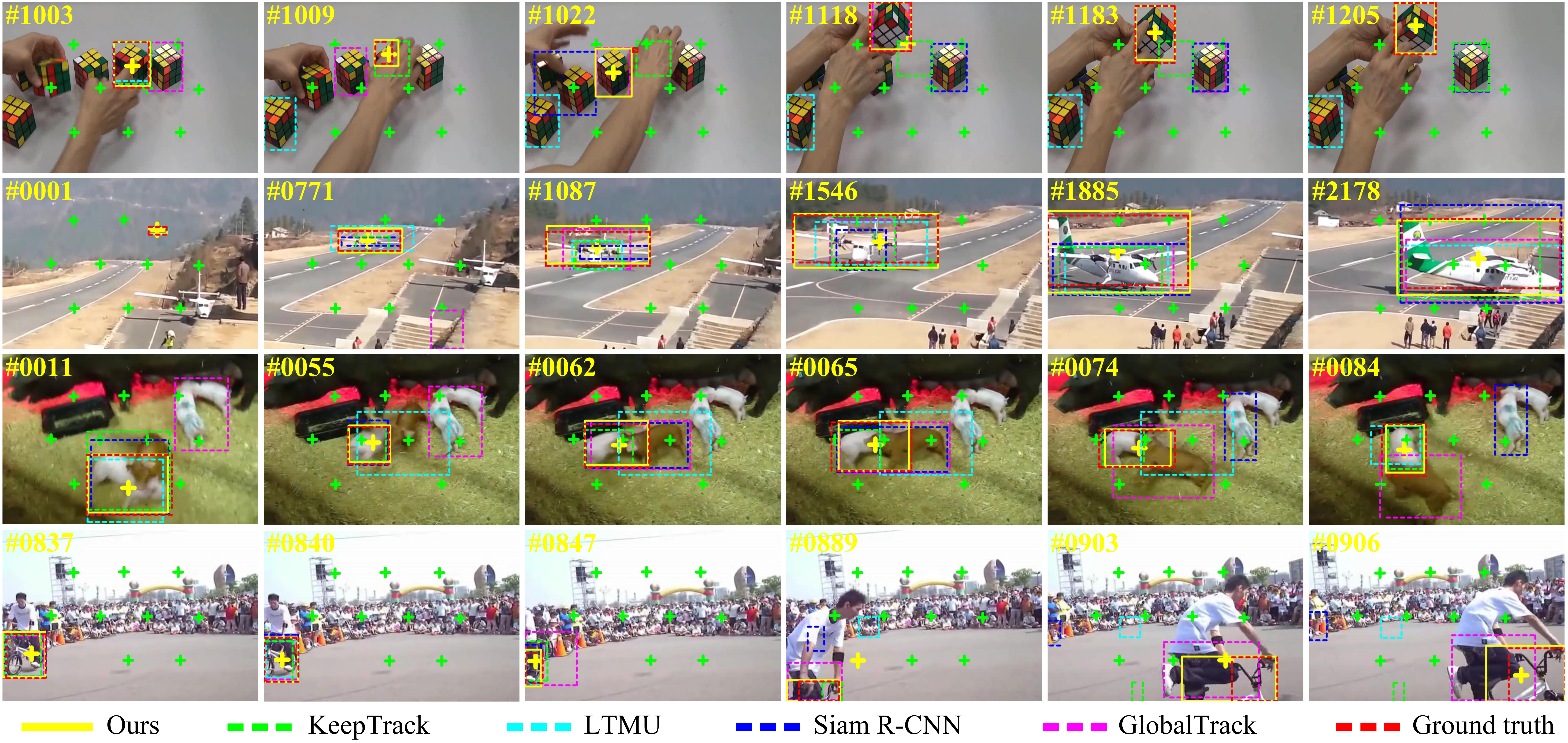}
\vspace{-2mm}
\caption{\textbf{Qualitative comparison on four challenging sequences.} From top to bottom, the main challenge factors are full occlusion, drastic appearance variation, distractor, and out-of-view, respectively. Our method is more robust than other long-term tracking methods.}
\label{Fig:Qualitative_comparison}
\vspace{-4mm}
\end{figure*}

\vspace{1mm}
\noindent\textbf{TrackingNet.} Besides the long-term tracking datasets, we evaluate our method on a short-term tracking dataset, TrackingNet~\cite{TrackingNet}. As shown in Table~\ref{Tab:TrackingNet}, our method performs favorably against the other transformer-based methods, TransT, STARK-ST50, and TrDiMP, by performance gains of 1.1\%, 1.2\%, and 4.1\% in AUC, respectively.

\vspace{1mm}
\noindent\textbf{Running Speed.} Table~\ref{Tab:LaSOTExtSub} reports the running speed of different trackers. Our method runs at 26 FPS, reaching the real-time speed, and runs faster than the long-term trackers including KeepTrack, LTMU, and GlobalTrack.

\vspace{-1mm}
\subsection{Qualitative Comparison}
\vspace{-1mm}
To obtain a qualitative comparison, we show the tracking results on four challenging sequences in Figure~\ref{Fig:Qualitative_comparison}, in which the main challenges are: full occlusion, appearance variation, distractor, and out-of-view, respectively. The qualitative comparison clearly shows that our method is more robust than other long-term trackers facing these challenges. For example, the first sequence shows the situation where the target is occluded and then reappears. When the target is almost fully occluded in the $1009^{th}$ frame, LTMU and GlobalTrack lose the target and drift to the distractor, while our method successes tracking the target. Although our method also drifts to a distractor when the target is fully occluded (the $1022^{th}$ frame), only our method successfully recovers to the real target when it reappears. KeepTrack predicts the target as absent since the $1009^{th}$ frame (still outputs a previously predicted bounding box). However, it fails to recover to the target when the target reappears and unfortunately drifts to a distractor at the $1205^{th}$ frame.
\vspace{-2mm}

\begin{figure}[t]
\centering
    \includegraphics[width=0.95\columnwidth]{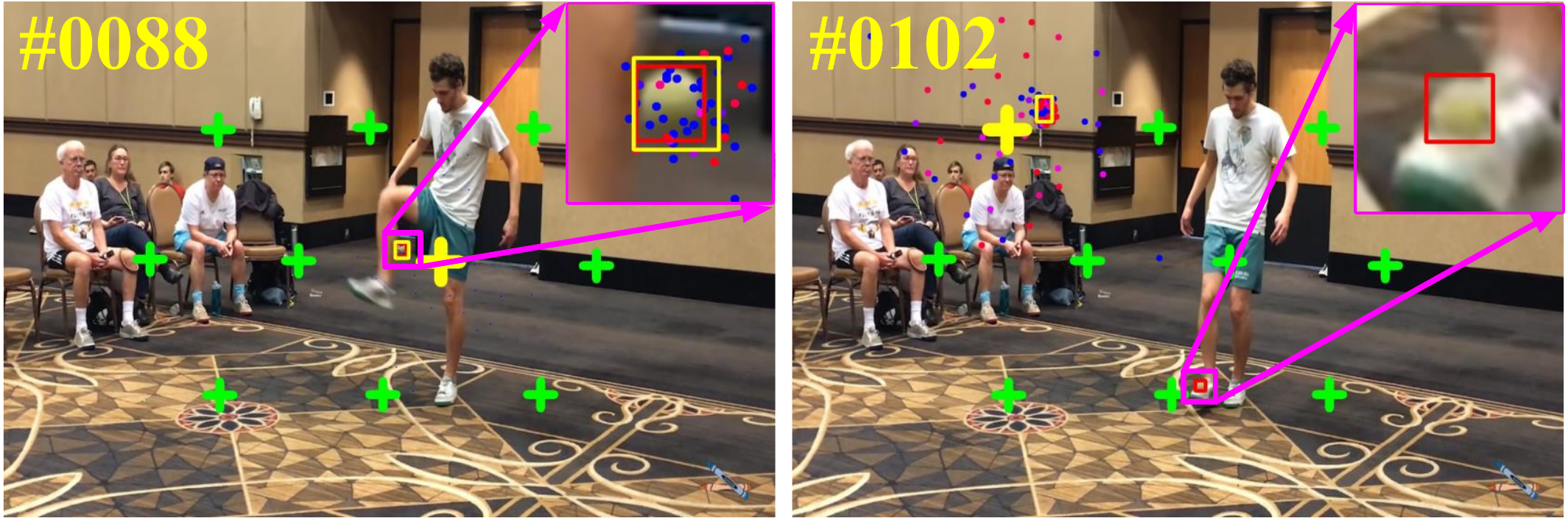}
\vspace{-1.2mm}    
\caption{\textbf{Failure case of our method on a \emph{footbag} sequence. } The targets in the two frames only occupy 0.35\textperthousand~(left) and 0.26\textperthousand~(right) of the image area, respectively. When the target becomes substantially small in the $102^{th}$ frame, whose information is overwhelmed by backgrounds, our method fails to locate it.}
\label{Fig:Failure_Case}
\vspace{-6mm}
\end{figure}

\subsection{Limitations}
\vspace{-1mm}
\label{Sec:Limitation}
Herein we discuss the limitations of our algorithm. Our local tracker is able to adjust its perception range by adaptively sampling the feature pixels from $\bm F_{t}$ to adapt to the objects of different sizes. However, this adaption mechanism does not work well for small objects. The reason is that small objects only occupy one or two feature pixels on $\bm F_{t}$, whose total stride is 16. Such limited target information tends to be overwhelmed by the backgrounds, making it arduous to locate the small objects. Figure~\ref{Fig:Failure_Case} shows the failure case of our method on a \emph{footbag} sequence. Besides, our method still cannot address the extremely challenging distractor issue well. As shown in the $1022^{th}$ frame in the first sequence in Figure~\ref{Fig:Qualitative_comparison}, our method locates the distractor when the target disappears, as most long-term trackers do.
\vspace{-1.75mm}

\section{Conclusion}
We have presented the global tracking algorithm via ensemble of local trackers, which can perform tracking in a global view while exploiting the temporal context. The smooth moving of the target can be handled by one single local tracker. If a local tracker loses the target due to discontinuous moving, another local tracker close to the target can take over the tracking to locate the target. Specifically, we design a deformable attention-based local tracker to simulate the local tracking mechanism within the global view. Further, we propose a temporal context transferring scheme to exploit the historical target appearances and locations for local tracking. The proposed method performs favorably against state-of-the-art trackers on six benchmarks.

\noindent\textbf{Acknowledgments.} This work was supported by the National Natural Science Foundation of China (U2013210, 62006060, and 62172126), the Shenzhen Stable Support Plan Fund for Universities (GXWD20201230155427003-20200824125730001), and the Shenzhen Research Council (JCYJ20210324120202006 and JCYJ20210324132212030).
{\small
\bibliographystyle{ieee_fullname}
\bibliography{track}
}

\end{document}